\documentclass[conference]{IEEEtran}
\IEEEoverridecommandlockouts
\usepackage{cite}
\usepackage{amsmath,amssymb,amsfonts}
\usepackage{algorithmic}
\usepackage{graphicx}
\usepackage{textcomp}
\usepackage{xcolor}
\usepackage{tikz}
\usepackage[nameinlink,capitalise]{cleveref}
\usetikzlibrary{arrows.meta, positioning, calc}

\def\BibTeX{{\rm B\kern-.05em{\sc i\kern-.025em b}\kern-.08em
    T\kern-.1667em\lower.7ex\hbox{E}\kern-.125emX}}
\begin{document}

\title{{LayerPipe2: Multistage Pipelining and Weight Recompute via Improved Exponential Moving Average for Training Neural Networks}\thanks{This research was supported in part by the National Science Foundation under grant number CCF-1954749.}}

\author{\IEEEauthorblockN{Nanda K. Unnikrishnan and Keshab K. Parhi, \textit{Life Fellow, IEEE}}
\IEEEauthorblockA{Dept. Electrical and Computer Engineering,\\
University of Minnesota Twin Cities\\
Minneapolis, MN 55455, USA}\\
Email: parhi@umn.edu}

\maketitle

\begin{abstract}
In our prior work, LayerPipe, we had introduced an approach to accelerate training of convolutional, fully connected, and spiking neural networks by overlapping forward and backward computation. However, despite empirical success, a principled understanding of how much gradient delay needs to be introduced at each layer to achieve desired level of pipelining was not addressed. This paper, LayerPipe2, fills that gap by formally deriving LayerPipe using variable delayed–gradient adaptation and retiming. We identify where delays may be legally inserted and show that the required amount of delay follows directly from the network structure: inner layers require fewer delays, while outer layers require longer delays. When pipelining is applied at every layer, each delay depends only on the number of remaining downstream stages; when layers are pipelined in groups, all layers in the group share the same assignment. These insights not only explain previously observed scheduling patterns but also expose an often-overlooked challenge: pipelining implicitly requires storage of historical weights. We overcome this storage bottleneck by developing a pipeline–aware moving average that reconstructs the required past states rather than storing them explicitly. This reduces memory cost without sacrificing the accuracy guarantees that makes pipelined learning viable. The result is a principled framework that illustrates how to construct LayerPipe architectures, predicts their delay requirements, and mitigates their storage burden, thereby enabling scalable pipelined training with controlled communication–computation tradeoffs.


\end{abstract}

\section{Introduction}
Deep neural networks (DNNs) have become central to modern machine learning systems~\cite{ojcas}, powering applications in computer vision~\cite{CV}, language modeling~\cite{LLM}, biomedical analysis~\cite{Bio}, and large-scale recommendation~\cite{DLRM}. As model sizes and dataset scales continue to grow, the training process is not an insignificant computational cost in deployed systems~\cite{training-cost}. Although accelerators such as GPUs~\cite{GPU}, TPUs~\cite{TPU}, and domain-specific NPUs~\cite{eyerissv2,SCV} offer increasing compute throughput, algorithmic restructuring of the training pipeline is essential to realize further speedups. In particular, reducing the strict sequential dependencies within backpropagation is key to improving utilization and throughput in multi-processor training environments~\cite{unnikrishnan2021layerpipe}.

Pipeline parallelism has become an important strategy toward this goal, enabling different segments of the model to process multiple inputs concurrently. However, pipelining backpropagation is fundamentally challenging due to the presence of nested feedback loops that involve weights, activations, and gradients~\cite{Ito, unnikrishnan2020gradientinterleaved,intergrad}. These loops restrict where delays may be introduced without altering functionality and convergence behavior. Existing pipeline-parallel frameworks address these challenges using heuristic approaches—such as weight stashing, activation stashing~\cite{harlap2018pipedream}, or microbatching—to manage inter-iteration state~\cite{huang2019gpipe, pipedream2}. While effective in practice, the underlying principles governing where delays can be inserted and how many delay cycles each layer requires have not been fully formalized.

\begin{figure}[t]
    \centering
    \includegraphics[width=\linewidth]{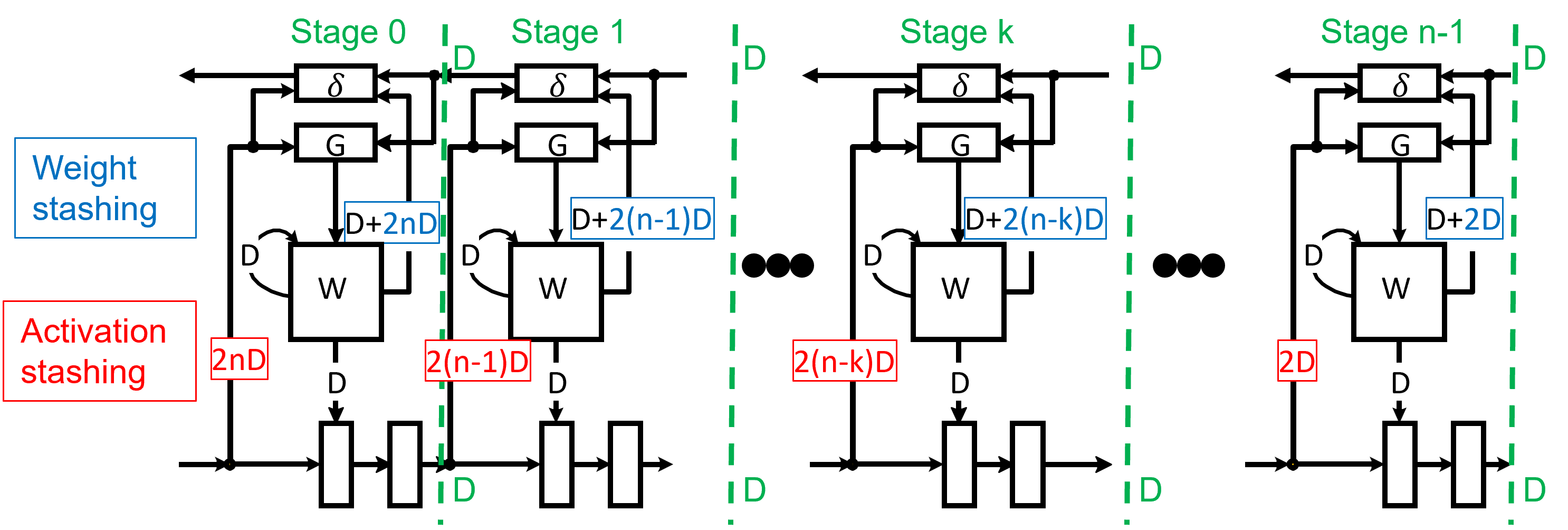}
    \caption{Dataflow graph of a modern pipeline parallel systems with stage-based pipelines and stashing for weights and activations.}
    \vspace{-0.7cm}
    \label{fig:pipeline}
\end{figure}

Our prior work on LayerPipe~\cite{unnikrishnan2021layerpipe} took an initial step toward resolving these limitations by analyzing the backpropagation data-flow graph through the lens of optimizations applied for DSP~\cite{parhi-vlsidsp}. In that paper, we showed how the backward pass can be decomposed into weight-gradient ($G$) and activation-gradient ($\delta$) computations, how feedforward cutsets enable the insertion of pipeline stages, and how retiming can be used to move delays across different parts of the graph. The earlier LayerPipe framework introduced these concepts through illustrative examples and demonstrated their effectiveness by constructing balanced schedules that achieved substantial speedup over existing pipeline-parallel methods. These examples suggest that a general and principled formulation is possible, capable of deriving the precise delay structure required for a given neural network architecture.

The present work further expands that foundation by developing a formal, general-purpose theory for pipelined backpropagation, grounded in feedforward cutset analysis\cite{cutset-analysis}, variable delayed gradient methodologies~\cite{delayed-gradient}, and retiming theory~\cite{retiming}. Building on the insights of the previous paper, we show how to systematically derive the number of gradient delays required at each layer, rather than treating delay patterns on a case-by-case basis. Specifically, we demonstrate that the required delay is proportional to the number of pipeline stages after a given layer, leading to fewer delayed gradients in the inner layers and more in the outer layers. For inter-layer pipelining in groups of $i$ layers, we prove that each layer in the group requires an identical amount of delay. These results formalize behaviors that were previously observed at a high level and establish mathematical expressions that generalize across arbitrary network depths.

By viewing backpropagation through the lens of data-flow graphs and adaptive filtering theory~\cite{adaptive-filtering}, we also show that activation stashing and weight stashing emerge naturally as consequences of retiming the underlying graph. This provides a theoretical basis for storage behaviors originally described intuitively and clarifies the connection between pipelining and delayed-gradient updates. In addition, this work extends LayerPipe to arbitrary multistage pipeline models, enabling deeper and more flexible pipeline structures than those demonstrated previously.
A deeper analysis of pipelining also highlights scalability challenges. As the number of pipeline stages increases, the amount of stashed weight state grows correspondingly. This can be seen in \cref{fig:pipeline}. To address this, we introduce a weight-prediction mechanism that reconstructs earlier weight versions using a pipeline-aware improved \textit{exponential moving average} (EMA). This method significantly decreases storage requirements while maintaining convergence quality. We validate the effectiveness of the proposed approach experimentally on ResNet-18 using CIFAR-100.

This paper makes the following key contributions:
\begin{itemize}
    \item  A formal derivation of pipelined backpropagation using cutset analysis and variable delayed gradient adaptation theory, generalizing concepts previously illustrated through specific examples.

\item A retiming-based formulation that produces explicit closed-form expressions for the number of gradient delays required at each layer, and explains activation and weight stashing as natural consequences of delay movement.

\item A general framework for multistage pipelining, extending LayerPipe~\cite{unnikrishnan2021layerpipe} to support deeper and more flexible pipeline layouts, including uniform pipelining across groups of layers.

\item A storage-efficient weight-prediction scheme based on an improved EMA that reconstructs historical weights with lower memory cost.

\item Experimental validation demonstrating that the proposed EMA improves convergence while substantially reducing weight-stashing requirements.

\end{itemize}
\section{Background and Related Work}

Training deep neural networks efficiently has driven a large body of parallelization research over the past decade. Early acceleration efforts focused on \emph{data parallelism}~\cite{dataparallelism}, wherein the full model is replicated on multiple devices and gradients are aggregated synchronously or asynchronously. While effective for moderate model sizes, the increasing depth and width of modern architectures expose the communication bottlenecks inherent in global gradient synchronization.

To address models that exceed per-device memory capacity, \emph{model parallelism}~\cite{modelparallelism} partitions the network across hardware units. While this allows scaling to larger models, it requires frequent exchange of intermediate variables across device boundaries, resulting in longer latency, load imbalance, and placement challenges~\cite{model_parallel_challenges}. These shortcomings led to an interest in \emph{pipeline parallelism}~\cite{huang2019gpipe, harlap2018pipedream}, which exploits the sequential nature of training to overlap computation across layers.

Pipeline parallelism was used in large-scale systems such as GPipe~\cite{huang2019gpipe} and PipeDream~\cite{harlap2018pipedream}. GPipe demonstrated substantial speedups by introducing microbatching to reduce idle periods between stages, while PipeDream further overlapped forward and backward computations by using stale weights and ``weight stashing'' to maintain consistency across pipeline iterations. These works showed that backpropagation tolerance to gradient staleness could be exploited to improve utilization. However, they also introduced heuristic mechanisms — including activation stashing, microbatch scheduling, and empirically tuned delay placements — that lack a universally applicable theoretical basis. As a consequence, generalizing such systems to new architectures or deeper pipelines remains difficult, because the conditions under which staleness does not impair convergence is not well understood~\cite{staleness_bounds}.

Delayed-gradient learning has long been studied in adaptive filtering, most notably in the delayed least mean square (DLMS) framework~\cite{delayed-gradient}. Delayed coefficients can preserve convergence under suitable step-size constraints. Classical DSP pipelines for recursive and adaptive systems introduced principled look-ahead and interleaving techniques to manage feedback-limited throughput and convergence~\cite{parhi1989-sla,shanbhag1993-rlalms,adaptive-filtering}. These ideas provide architectural context for delayed-gradient pipelining in backpropagation. Although these insights align with the staleness toleration implicitly exploited by PipeDream, they were largely disconnected from how pipeline scheduling is practiced in deep learning.

Recent work began bridging this gap by treating backpropagation explicitly as a dataflow graph~\cite{unnikrishnan2021layerpipe}. The LayerPipe framework showed, through concrete examples, that layer boundaries form feedback cutsets where gradient delays may be inserted without violating correctness via variable delayed gradient adaptation. By decomposing the backward pass into weight- and activation-gradient components, LayerPipe demonstrated intra-layer and inter-layer pipeline opportunities consistent with observed scheduling practices. Similar optimizations appear in alternative learning paradigms such as spiking neural networks, where LayerPipe-inspired delayed-update formulations have been explored to reduce simulation latency and enable asynchronous synaptic learning~\cite{layerpipe_spiking}. These results suggested that retiming theory — traditionally used in hardware pipeline construction~\cite{retiming} — might offer a principled foundation for understanding and designing training pipelines. However, the initial formulation left open questions around generalization: how many delay elements are required at each layer, how pipeline depth influences staleness, and why phenomena such as weight stashing emerge structurally rather than heuristically.

Parallel to developments in scalable training, research on \emph{weight averaging} investigated the use of exponential moving averages (EMA) to improve robustness, calibration, and generalization of trained models~\cite{moralesbrotons2024ema}. These studies focused on improving solution quality, showing that EMA-smoothed weights outperform last-iterate weights in several settings. In contrast to such uses of EMA as a model “teacher,” pipeline execution demands reconstruction of specific historical weight states so that delayed gradients may be applied correctly, and this requirement has received comparatively little formal treatment. Existing systems either maintain multiple stored weight versions (as in PipeDream)~\cite{harlap2018pipedream} or constrain pipeline depth to limit storage~\cite{pipedream2}, without offering a model-based mechanism to recover older weights.

Taken together, these observations point to two outstanding challenges. First, pipeline parallelism lacks a general theoretical foundation capable of predicting where delays may be inserted and how many are required for correctness across arbitrary networks. Second, execution-driven staleness introduces non-trivial storage costs that call for principled reconstruction mechanisms rather than direct state duplication. These gaps motivate the framework developed in this paper.

\begin{figure}[t]
    \centering
    \includegraphics[width=\linewidth]{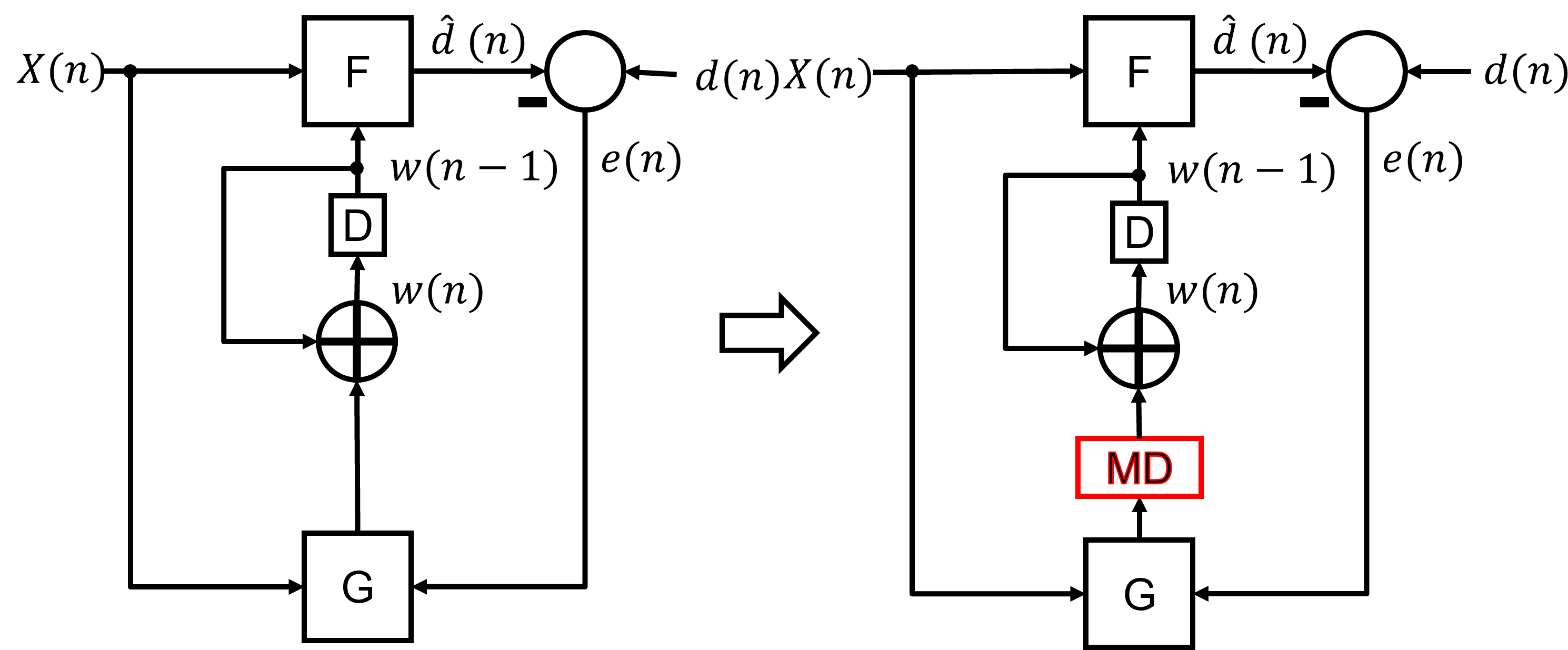}
    \caption{
        Conceptual delayed least–mean–square (DLMS) adaptation, in which an $M$-sample delay is introduced in the coefficient update path. 
        This abstraction models the effect of stale gradients within iterative learning systems and provides a historical analogue to delayed gradient application in pipelined training.
    }
    \label{fig:delayed_lms}
\end{figure}

\section{Proposed Method}

This section formalizes delayed–gradient pipelining using retiming theory. We begin by establishing the structural properties that make delay insertion possible, develop a graph–retiming formulation that produces pipeline stages, then extend the construction to multistage settings and a storage–efficient weight reconstruction mechanism.

\subsection{Delayed Gradient Foundations}

The feasibility of using stale gradients for optimization traces back to delayed least mean square (DLMS) theory, which established that delayed coefficient updates converge for slowly-varying processes and under appropriate stability conditions. As illustrated in \cref{fig:delayed_lms}, this effect is modeled by inserting an $M$-sample delay between the error computation and parameter update, resulting in delayed gradients being applied to earlier weight states. This viewpoint directly parallels pipelined neural training: delayed gradients arising from inter-stage overlap behave similarly to the DLMS delayed update, suggesting that controlled delay can be tolerated without violating correctness.

A useful observation is that input and output boundaries in standard neural networks form \emph{feedforward cutsets}, where signals flow only forward. Delay elements inserted at such cutsets preserve functional equivalence. While earlier work illustrated this principle through example networks, here we express the requirement in closed form.

Let $S(l)$ denote the number of pipeline stages after layer $l$. Then the number of delay elements required to align gradient arrival with the weight–update operation is
\begin{equation}
    \mathrm{Delay}(l) = 2 S(l),
\end{equation}
a rule that generalizes example–driven constructions and applies uniformly across architectures.

\begin{figure*}[t]
    \centering
    \includegraphics[width=0.85\linewidth]{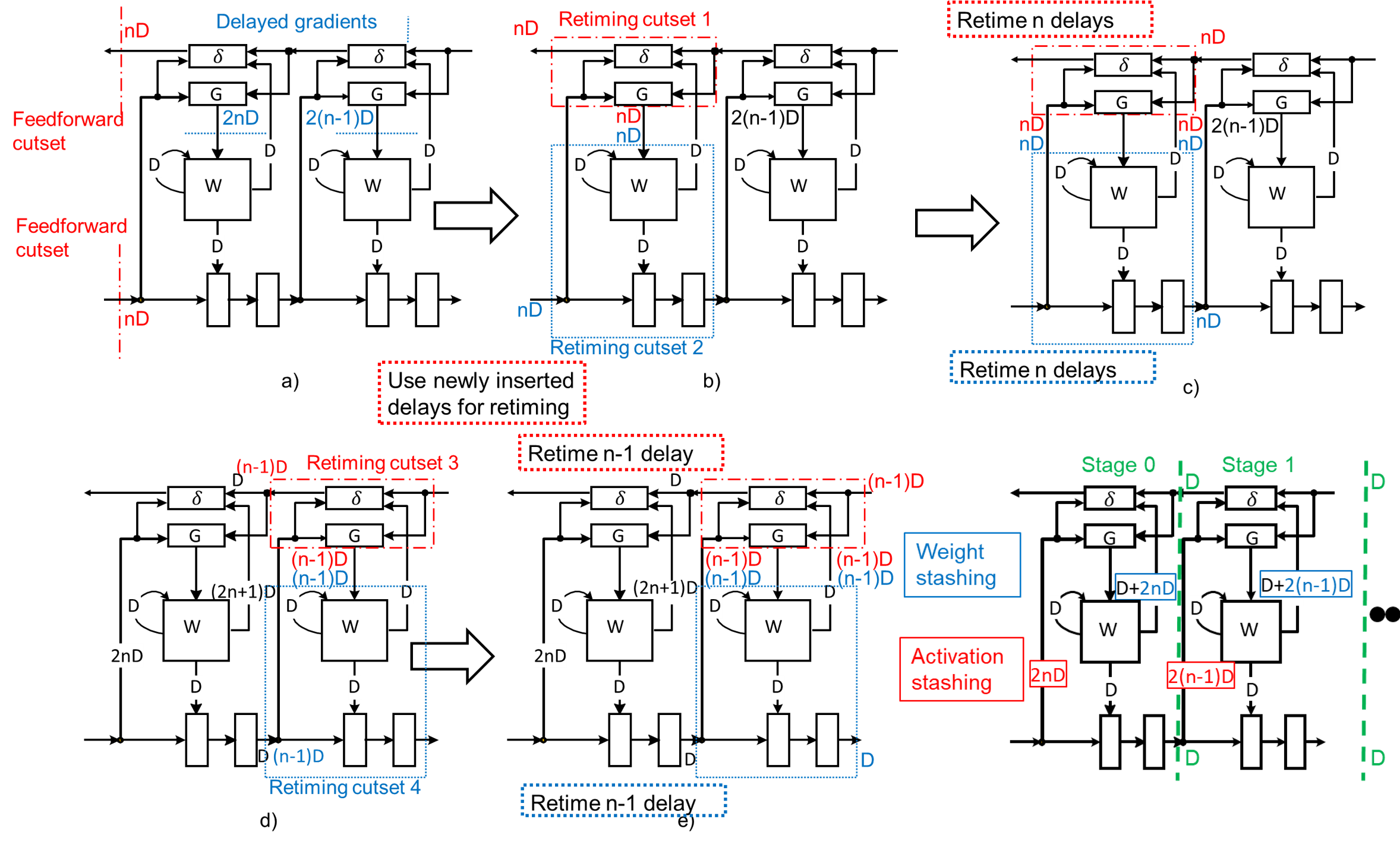}
    \caption{
        Retiming-based derivation of LayerPipe pipelining. 
        Delays are first inserted at feedforward cutsets and weight-update edges, then progressively retimed across cutsets so that delay elements accumulate at layer boundaries. 
        The resulting temporal separation forms explicit pipeline stages and reveals activation and weight stashing as structural consequences of delay motion. 
        The delay placements match the analytical rule $\mathrm{Delay}(l)=2S(l)$ derived in the text.
    }
    \label{fig:layerpipe_derivation}
\end{figure*}

\subsection{Retiming-Based Derivation of Pipelined Backpropagation}

Backpropagation admits a dataflow representation in which activations propagate forward and gradients propagate backward. We construct a pipelined architecture by inserting delays at locations where delayed updates are valid and then relocating these delays so they align with desired stage boundaries. \cref{fig:layerpipe_derivation} illustrates the evolution of these delays into explicit pipeline stages.

\begin{enumerate}
    \item \textbf{Delay insertion at feedforward cutsets:}
    Two feedforward cutsets exist at the network input and output. These admit delay
    insertion without affecting correctness. For an $n$-stage pipeline, each cutset
    receives a delay of $nD$, annotated on the forward edges in
    \cref{fig:layerpipe_derivation}.

    \item \textbf{Delay insertion on gradient feedback edges (DLMS–inspired):}
    The gradient-to-weight update path forms a feedback loop. DLMS demonstrates that
    delayed application remains valid, so delays may also be inserted here. For a layer
    with $n$ downstream stages, an additional round-trip delay of $2nD$ is assigned to
    the gradient update edge, and analogously $2(n-1)D, 2(n-2)D,\dots$ for deeper layers.

\item \textbf{Retiming across cutsets:}
Delay relocation is carried out using two retiming cutsets. The first retiming cutset operates on the backward stage. It shifts $nD$ delay units from all outward edges of the backward domain—including activation gradients to the preceding layer and the gradient–weight feedback edge—onto the corresponding inward backward edges: activation-to-gradient paths, weight-to-gradient paths, and the backward input from the following layer.

The second retiming cutset operates on the forward stage. It shifts $nD$ delay units from all inward forward edges—including input from the previous layer and the forward traversal of the gradient–weight feedback—onto the outward forward edges: activation-to-gradient paths, weight-to-gradient paths, and forward activation to the next layer.

These two cutset operations together explain why the gradient–weight feedback edge carries a $2nD$ assignment: one $nD$ term participates in the backward retiming pass and one $nD$ in the forward pass, matching the round-trip traversal reflected in~\cref{fig:layerpipe_derivation}.

\item \textbf{Recursive delay compaction:}
After the two retiming passes, $nD$ delay units reside on the edges associated with the current layer. One of these delay units is intentionally left in place; it represents the pipeline stage boundary that will exist at this location. The retiming procedure is then reapplied to the remaining $(n-1)D$ delay units, moving them through the same backward and forward cutsets. Again, one delay unit is left in place to mark the next stage boundary.

This process repeats until all delay units are exhausted. Because each iteration leaves one delay at its boundary, the number of delays processed at each step decreases from $nD$ to $(n-1)D$, $(n-2)D$, and so on. As a result, the delay associated with a layer is directly proportional to the number of pipeline stages that lie after it. Layers closer to the output retain more delay and therefore participate in deeper round-trip traversal, which is exactly the pattern visible in \cref{fig:layerpipe_derivation}.
\end{enumerate}

\cref{fig:layerpipe_derivation} visualizes this evolution: delay elements introduced at cutsets migrate inward under successive retiming operations until they accumulate at stage boundaries. States displaced by retiming must remain available when delayed gradients return, which manifests as activation and weight stashing. Thus, the pipeline structure is not imposed externally; it emerges as a direct consequence of how delays propagate through the backpropagation graph.

The resulting structure is a pipeline partitioning derived formally rather than heuristically: delays determine the stage boundaries, not the reverse.

\subsection{Generalizing to Arbitrary Pipeline Partitions}

The previous derivation illustrated the special case in which each layer forms a distinct pipeline stage. However, the same delay-construction and retiming principles extend to any partitioning of layers into stages. If a group of $i$ consecutive layers is assigned to a single stage, then every layer in that group shares the same downstream stage count, and therefore carries the same delay requirement. The retiming process does not operate layer-by-layer, but rather across the entire grouped region, treating it as a single logical unit.

\begin{figure}[t]
    \centering
    \includegraphics[width=\linewidth]{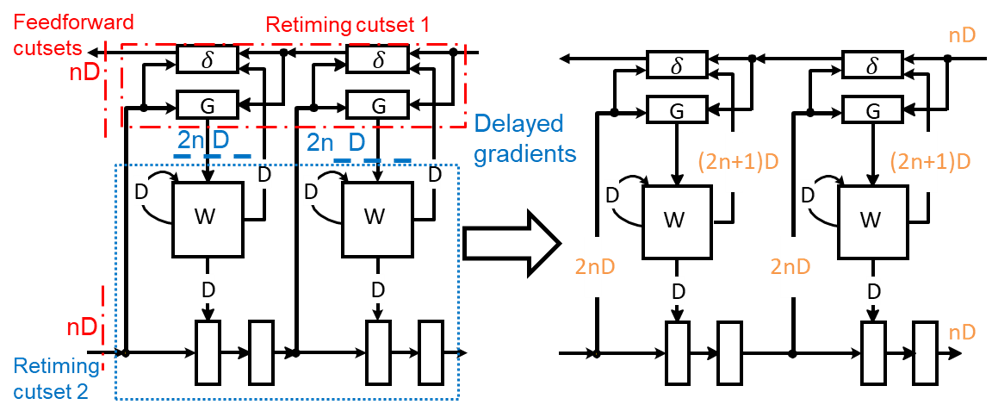}
    \caption{
        Illustration of retiming applied over a grouped two-layer stage.
        Delays are inserted at feedforward cutsets and gradient feedback edges,
        then redistributed using backward and forward retiming cutsets.
        One delay element is left at each downstream stage boundary, yielding a structure
        in which both layers share identical delay requirements. This demonstrates that
        delay placement is determined by the number of pipeline stages that follow the
        grouped region, not by the number of layers it contains.
    }
    \label{fig:multistage}
\end{figure}

\cref{fig:multistage} illustrates this over two layers grouped into one pipeline stage. Feedforward cutsets still inject $nD$ delays on forward paths, while delayed-gradient insertion provides $2nD, 2(n-1)D, \ldots$ on the gradient–weight feedback edges. Retiming cutset~1 shifts delays from backward outward edges to their corresponding backward inward edges across the entire two-layer region, and retiming cutset~2 performs the analogous movement in the forward direction. The compaction step again leaves one delay element per downstream stage boundary.

The resulting structure demonstrates that delay accumulation and redistribution do not depend on per-layer staging; instead, they depend only on how many pipeline boundaries exist downstream of the grouped region. Accordingly, the closed-form rule for delay placement remains unchanged: the delay associated with a region is proportional to the number of stages after that region, independent of how many layers the region spans.

\subsection{Storage Optimization via Weight Prediction}

The delay construction and retiming steps above imply that each layer must access historical weights when delayed gradients arrive. In a direct implementation, this requirement is satisfied by \emph{weight stashing}: for every layer and every pipeline stage, a separate copy of the parameter vector is stored. For a network with $L$ layers and a maximum of $n$ stages after a given layer, the memory cost scales as $\mathcal{O}(Ln)$ in addition to activation stashing. As models and pipeline depths grow, this quickly becomes the dominant storage term.

We instead reconstruct the historical weight from the current weight and an estimate of the intervening gradients. Let $W_\ell(t)$ denote the weight vector of layer $\ell$ at iteration $t$, $G_\ell(t)$ the corresponding gradient, and $\alpha$ the learning rate of a first-order optimizer such as SGD. For a layer whose gradients are applied after a round-trip delay of $(2n+1)$ iterations, the SGD update can be written as
\begin{equation}
    W_\ell(t)
    = W_\ell\bigl(t-(2n+1)\bigr)
      - \alpha \sum_{i=0}^{2n+1} G_\ell(t-i).
    \label{eq:sgd-forward}
\end{equation}
Rearranging the SGD update for a layer $\ell$ with round–trip delay $(2n+1)$ gives an exact expression for the historical weight
\begin{equation}
    W_\ell\bigl(t-(2n+1)\bigr)
    = W_\ell(t) + \alpha \sum_{i=0}^{2n+1} G_\ell(t-i),
    \label{eq:historic-weight-exact}
\end{equation}
The storage problem is now shifted to the finite gradient sum on the right-hand side.

To avoid storing past gradients explicitly, we introduce an averaged gradient. To obtain a moving average that approximates the finite gradient sum in
\cref{eq:historic-weight-exact}, we define an averaged gradient sequence
\begin{equation}
    \bar{G}(n) = \frac{1}{n+1}\sum_{i=0}^{n} G(i),
    \label{eq:avgdef2}
\end{equation}
where $\bar{G}(n)$ denotes the average of the most recent $(n+1)$ gradients. The quantity $\bar{G}(n-1)$ is defined analogously over $n$ samples. A direct recurrence for $\bar{G}(n)$ follows immediately:
\begin{align}
    (n+1)\bar{G}(n)
        &= \sum_{i=0}^{n} G(i)
         = G(n) + \sum_{i=0}^{n-1} G(i) \\
        &= G(n) + n\,\bar{G}(n-1),
    \label{eq:avgexpand}
\end{align}
which yields
\begin{equation}
    \bar{G}(n)
        = \frac{n}{n+1}\,\bar{G}(n-1)
          + \frac{1}{n+1}\,G(n).
    \label{eq:avgrecfinal}
\end{equation}

\Cref{eq:avgrecfinal} has the structure of an exponential moving average (EMA) \cite{moralesbrotons2024ema},
\[
    \bar{G}(n) = \beta(n)\,\bar{G}(n-1)
                 + \bigl(1-\beta(n)\bigr)G(n),
\]
with the decay term, $\beta(n)$, obtained analytically as
\begin{equation}
    \beta(n) = \frac{n}{n+1},
    \qquad
    1-\beta(n) = \frac{1}{n+1}.
    \label{eq:betacondensed}
\end{equation}

Thus, the EMA coefficient needed to reconstruct a window of length $(n+1)$  follows directly from the averaging constraint. Matching the window length $(n+1)$ to the pipeline delay $(2n+1)$ provides the required averaged gradient value without storing multiple historical parameter states. Substituting this expression into \cref{eq:historic-weight-exact} gives the equation for the approximated, $\widehat{W}_\ell\bigl(t-(2n+1)\bigr)$, as
\begin{equation}
    \widehat{W}_\ell\bigl(t-(2n+1)\bigr)
        = W_\ell(t) + \alpha\,(2n+1)\,\bar{G}_\ell(n),
\end{equation}
where $\bar{G}_\ell(n)$ is updated online via \cref{eq:avgrecfinal}. This establishes a closed-form, pipeline-aware reconstruction rule that eliminates explicit weight stashing while preserving delayed-update correctness.
\section{Experimental Results}

Where the earlier LayerPipe study emphasized throughput gains from intra-- and inter--layer pipelining, the focus here is complementary. Our goal is to empirically assess how pipelining interacts with optimization dynamics, and whether accuracy can be preserved without explicit weight stashing.

\subsection{Experimental Setup}

All experiments were conducted on a workstation equipped with an Intel 12th Gen Core i7--12700H CPU (2.30~GHz base frequency), 16~GB RAM, and an NVIDIA GeForce RTX~3070\,Ti GPU using a 64--bit operating environment with CUDA/cuDNN acceleration.

We evaluate on CIFAR--100 using ResNet--18 as the base architecture. Training proceeds for 50 epochs with a batch size of 128. For pipelined configurations, the computation graph is partitioned into eight forward--backward scheduling units, allowing concurrent advancement across pipeline stages. Optimization uses stochastic gradient descent with momentum and weight decay. The initial learning rate is 0.1 and decays smoothly via cosine annealing over the full training horizon.

Pipeline--aware weight prediction is activated following a 2--epoch warm--up period, during which exponential moving averages stabilize before being used to reconstruct historical weights.

\subsection{Baselines}

We compare against the following staleness--handling strategies:

\begin{itemize}
    \item \textbf{Sequential backpropagation}: standard, nonpipelined training.
    \item \textbf{Weight stashing}: pipelined execution that stores full historical weight states (baseline pipeline).
    \item \textbf{Latest--weight approximation}: applying delayed gradients to current weights.
    \item \textbf{Fixed--decay EMA}: historical weights approximated with a conventional moving average, independent of delay depth. For this experiment, a $\beta$ of 0.9 is chosen.
    \item \textbf{Proposed pipeline--aware EMA}: delay--conditioned weight reconstruction.
\end{itemize}

These represent increasingly lightweight alternatives to exact stashing, allowing us to quantify accuracy degradation and potential recovery.

\subsection{Convergence Behavior}

\begin{figure}[t]
    \centering
    \includegraphics[width=\linewidth]{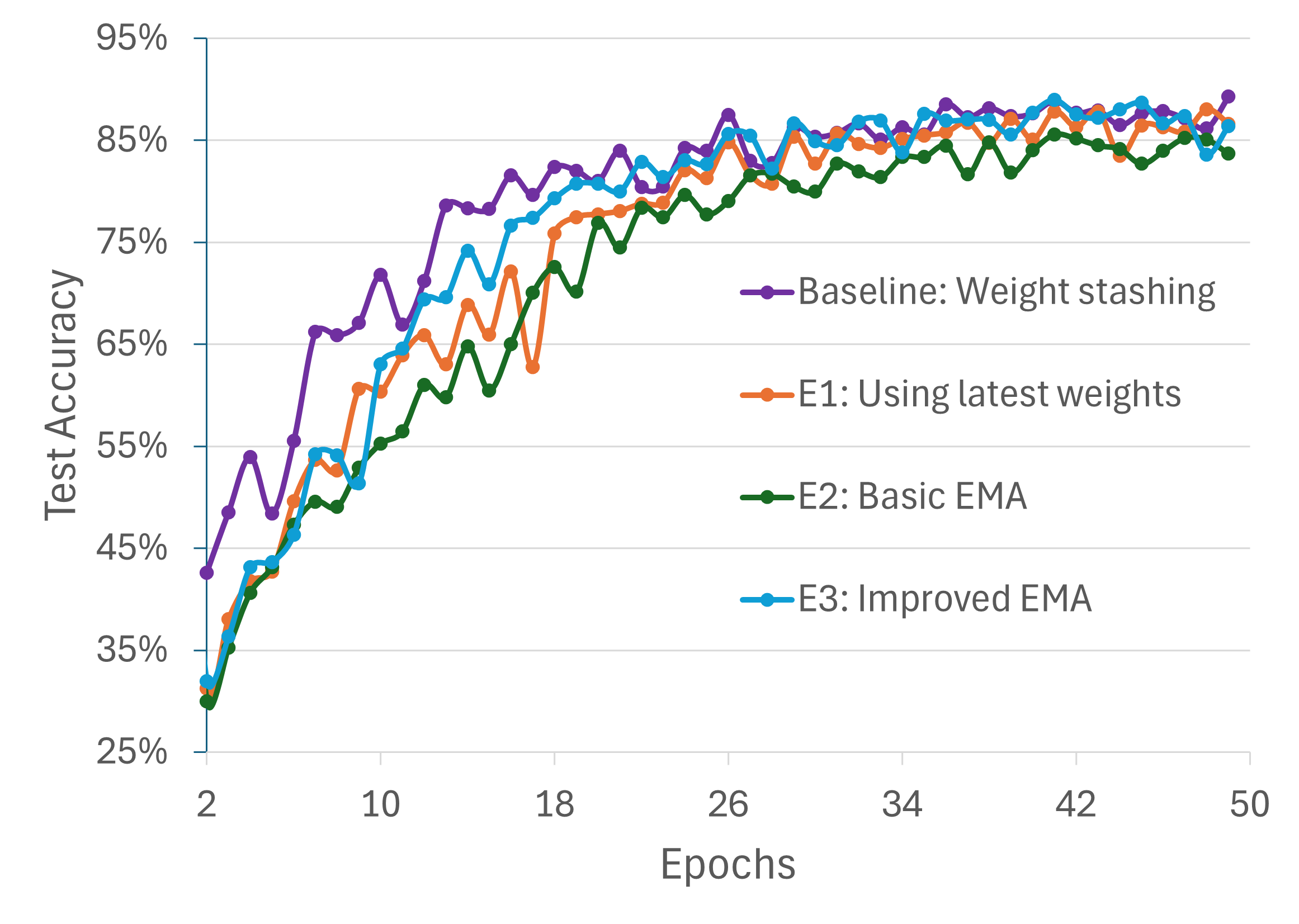}
    \caption{Test accuracy over fifty epochs on CIFAR--100 for four weight–handling strategies under pipelined training. Explicit weight stashing (baseline) produces stable convergence. Using the latest weights degrades performance, and a fixed EMA only partially recovers accuracy. The proposed pipeline-aware EMA reconstructs historical weights sufficiently accurately to match the baseline, but without storing multiple weight versions.}
    \label{fig:convergence_results}
\end{figure}

\cref{fig:convergence_results} reports test accuracy over 50 epochs for all methods. As expected, the stashing baseline yields stable convergence, since each gradient is applied to the weight version it was computed against. By contrast, directly applying delayed gradients to the latest weights results in slower and more erratic convergence, indicating that mismatched update application adversely affects optimization.

A conventional EMA partially mitigates this effect, smoothing weight evolution and reducing instability, yet its final accuracy consistently trails the baseline. In contrast, the proposed pipeline--aware EMA closely tracks the stashing baseline after its warm--up period. The convergence trajectories of the two curves quickly align, validating that delay--conditioned prediction recovers the correct update behavior without storing multiple weight versions.

\subsection{Interpretation}

Although the evaluation centers on a single benchmark, the result highlights a fundamental insight. The detrimental effects of pipeline--induced staleness are not inherent to pipelining; they arise from misaligned weight states and can be corrected without storage duplication. This reinforces the theoretical claim that stashing emerges structurally from delay retiming, while also demonstrating that it can be replaced computationally through informed reconstruction.

Accordingly, the proposed method complements prior LayerPipe results: previous work established that pipelining exposes latent parallelism and improves utilization, whereas the present experiment shows that its convergence penalties can be mitigated efficiently. Together, these findings strengthen the practicality of deep training pipelines by coupling performance gains with accuracy preservation.

\section{Conclusion}

This work presented a formal framework for constructing pipelined backpropagation architectures by combining delayed-gradient theory with retiming principles. Building on the pipeline scheduling insights of the earlier LayerPipe study, which demonstrated the throughput benefits of exploiting intra-- and inter--layer concurrency, we developed the foundational rules that govern where delays may be inserted and how many are required for correctness. The derivation shows that activation and weight stashing arise directly from delay movement within the dataflow graph, rather than from heuristic implementation choices.

The generalization to multistage pipelining enables arbitrary partitioning of network layers and establishes a closed-form expression for delay assignment across stages. This theoretical groundwork provides a means of designing deeper pipelines and reasoning about their communication--computation tradeoffs. A second contribution addresses a limitation of prior pipeline systems: the cost of storing historical weight states. We introduced a pipeline--aware exponential moving average predictor that reconstructs older weights using current parameters and a tuned gradient accumulator, reducing memory overhead from $O(LS)$ to $O(L)$ while maintaining accuracy.

Experiments confirm that the proposed delay model yields functionally correct execution schedules, that multistage pipelining improves concurrency as expected from earlier LayerPipe results, and that prediction-driven staleness compensation recovers convergence performance otherwise lost in naive pipelined training. In this sense, the contributions are complementary: prior work established that pipelining exposes latent parallelism, while the present results show how to preserve training fidelity as pipeline depth increases.

Future extensions include applying the framework to large transformer architectures, incorporating adaptive delay selection into the training process, and exploring interactions between delayed-gradient theory and optimizer design. More broadly, the retiming interpretation suggests connections between neural training dynamics and classical digital system pipelining, opening opportunities for co-designed algorithm–architecture pipelines informed by formal delay analysis.

\bibliographystyle{ieeetr}

\bibliography{references}
\end{document}